\documentclass{article}

\PassOptionsToPackage{numbers, compress}{natbib}
\usepackage[nonatbib,preprint]{neurips_2022}

\usepackage[utf8]{inputenc} % allow utf-8 input
\usepackage[T1]{fontenc}    % use 8-bit T1 fonts
\usepackage{hyperref}       % hyperlinks
\usepackage{url}            % simple URL typesetting
\usepackage{booktabs}       % professional-quality tables
\usepackage{amsfonts}       % blackboard math symbols
\usepackage{nicefrac}       % compact symbols for 1/2, etc.
\usepackage{microtype}      % microtypography
\usepackage{xcolor}         % colors
\usepackage{microtype}
\usepackage{setspace}
\usepackage{multirow}
\usepackage{multicol}
\usepackage{enumitem}
\usepackage{threeparttable}
\usepackage{xspace}
\usepackage{amsmath}
\usepackage{amsthm}
\usepackage{amssymb}
\usepackage{booktabs}
\usepackage{adjustbox}
\usepackage{colortbl}
\usepackage{xcolor}
\usepackage{diagbox}    
\usepackage{algorithmic}
\usepackage{algorithm}
\usepackage{wrapfig}

\setlist[itemize]{leftmargin=5.5mm}
\newcommand{\method}{HIRL}
\definecolor{b}{RGB}{255,0,0}
\definecolor{w}{RGB}{255,80,80}
\definecolor{l}{RGB}{0,0,255}
\definecolor{s}{RGB}{120,120,120}

\title{HIRL: A General Framework for \\ Hierarchical Image Representation Learning}

\author{
	\textbf{Minghao Xu}\textsuperscript{\rm 1,3$\,*$} \quad
	\textbf{Yuanfan Guo}\textsuperscript{\rm 2$\,*$} \quad
	\textbf{Xuanyu Zhu}\textsuperscript{\rm 4} \quad
	\textbf{Jiawen Li}\textsuperscript{\rm 4} \quad
	\textbf{Zhenbang Sun}\textsuperscript{\rm 4} \\
	\textbf{Jian Tang}\textsuperscript{\rm 1,5,6} \quad
	\textbf{Yi Xu}\textsuperscript{\rm 2$\,\dagger$} \quad
	\textbf{Bingbing Ni}\textsuperscript{\rm 2$\,\dagger$} \\
	\textsuperscript{\rm *}{\small equal contribution} \quad
	\textsuperscript{\rm $\dagger$}{\small corresponding author} \\
	\textsuperscript{\rm 1}Mila - Qu\'{e}bec AI Institute \\
	\textsuperscript{\rm 2}MoE Key Lab of Artificial Intelligence, AI Institute, Shanghai Jiao Tong University \\
	\textsuperscript{\rm 3}University of Montr\'{e}al \quad
	\textsuperscript{\rm 4}ByteDance \quad
	\textsuperscript{\rm 5}HEC Montr\'{e}al \quad 
	\textsuperscript{\rm 6}CIFAR AI Research Chair \\
	{\small \textbf{contacts:} minghao.xu@umontreal.ca, $<$gyfastas, xuyi, nibingbing$>$@sjtu.edu.cn}
}

\begin{document}

\maketitle

%%%%%%%%%%%%%%%%%%%%%%%%%%%%%%%%%%%%%%%%%%%%%%%%%%%%%%%%%%%%

%%%%%%%%%%%%%%%%%%%%%%%%%%%%%%%%%%%%%%%%%%%%%%%%%%%%%%%%%%%%

\begin{abstract}

Learning self-supervised image representations has been broadly studied to boost various visual understanding tasks. Existing methods typically learn a single level of image semantics like pairwise semantic similarity or image clustering patterns. However, these methods can hardly capture multiple levels of semantic information that naturally exists in an image dataset, \emph{e.g.}, the semantic hierarchy of ``Persian cat $\rightarrow$ cat $\rightarrow$ mammal'' encoded in an image database for species. It is thus unknown whether an arbitrary image self-supervised learning (SSL) approach can benefit from learning such hierarchical semantics. To answer this question, we propose a general framework for \textbf{H}ierarchical \textbf{I}mage \textbf{R}epresentation \textbf{L}earning ({\method}). This framework aims to learn multiple semantic representations for each image, and these representations are structured to encode image semantics from fine-grained to coarse-grained. Based on a probabilistic factorization, {\method} learns the most fine-grained semantics by an off-the-shelf image SSL approach and learns multiple coarse-grained semantics by a novel \textbf{semantic path discrimination} scheme. 
We adopt six representative image SSL methods as baselines and study how they perform under {\method}. By rigorous fair comparison, performance gain is observed on all the six methods for diverse downstream tasks, which, for the first time, verifies the general effectiveness of learning hierarchical image semantics. All source code and model weights are available at \url{https://github.com/hirl-team/HIRL}.
% By benchmarking six representative image SSL methods on a common platform and further adapting them into the {\method} framework, performance gain is observed on all the six methods for diverse downstream tasks, which, for the first time, verifies the general effectiveness of learning hierarchical image semantics. 

\end{abstract}

%%%%%%%%%%%%%%%%%%%%%%%%%%%%%%%%%%%%%%%%%%%%%%%%%%%%%%%%%%%%
%%%%%%%%%%%%%%%%%%%%%%%%%%%%%%%%%%%%%%%%%%%%%%%%%%%%%%%%%%%%

\vspace{-3mm}
\section{Introduction} \label{sec:intro}
\vspace{-1mm}

Self-supervised image representation learning has been enthusiastically studied in recent years. It focuses on acquiring informative and interpretable feature representations for unlabeled images, where the learning signal totally comes from the data itself. To attain this goal, a variety of self-supervised learning (SSL) techniques have been developed to capture different levels of semantics underlying the raw images, \emph{e.g.}, capturing the semantic similarity between image pairs~\cite{moco,simclr,moco_v2,moco_v3} and capturing the semantics of image clusters~\cite{deepcluster,swav,pcl,dino}. The image representations learned by these methods have been proven to be effective on various downstream tasks, and the performance gap between SSL and fully-supervised learning is continually narrowing.

However, for modeling a large-scale image dataset, it is always not enough to capture a single level of semantic information, because of the fact that such a dataset commonly contains multiple semantic hierarchies. For example, in a dataset of diverse species, an image of \emph{Persian cat} also owns a more coarse-grained semantics of \emph{cat} and an even more coarse-grained semantics of \emph{mammal}. Encoding such hierarchical semantic information (\emph{e.g.}, Persian cat $\rightarrow$ cat $\rightarrow$ mammal) in image representations can broadly benefit different downstream tasks, \emph{e.g.}, image classification with different label granularity. There are some recent works~\cite{deepercluster,pcl,hcsc} aiming to learn such hierarchical semantic representations, and some performance gain has been verified by incorporating such a scheme. However, there still lacks a systematic study on \textbf{whether any off-the-shelf image SSL method can benefit from learning multiple levels of semantic information}.
\vspace{-0.4mm}

To answer this question, in this work, we propose a general framework for \textbf{H}ierarchical \textbf{I}mage \textbf{R}epresentation \textbf{L}earning (\emph{\method}). This framework seeks to learn multiple representations for each image, and these representations encode the image's different levels of semantics from fine-grained to coarse-grained. To set up a starting point of this semantic hierarchy, we employ an arbitrary off-the-shelf image SSL method to learn the representation of the most fine-grained semantics. On such basis, we propose a novel \textbf{semantic path discrimination} (\emph{SPD}) approach to learn the representations of more coarse-grained semantics. This approach first identifies some representative embeddings (\emph{i.e.}, prototypes) of underlying semantic clusters on different semantic levels. For each image, it then retrieves a positive path composed of the prototypes with the same hierarchical semantics as the image and retrieves multiple negative paths composed of the prototypes that encode different semantics with the image. The objective of SPD is to maximize the similarity between the image's hierarchical representations and the positive path while minimize the similarity between the hierarchical representations and the negative paths. 
In this way, the {\method} framework can incorporate the objective of hierarchical semantic representation learning into various off-the-shelf image SSL methods, which strengthens their capability of modeling image semantics. 
% In this way, the {\method} framework combines the objective of an arbitrary off-the-shelf SSL method and the objective for hierarchical image representation learning, which enables to evaluate the extra effect of learning multiple levels of semantic information.
\vspace{-0.4mm}

For fair comparison, we first re-implement three representative CNN based SSL algorithms and three representative Vision Transformer based SSL algorithms under a common codebase. We then adapt these six image SSL methods to the {\method} framework and compare their performance before and after enabling hierarchical semantic modeling. The experimental results on standard downstream evaluation tasks, \emph{e.g.}, KNN evaluation, linear classification and fine-tuning on ImageNet~\cite{imagenet}, verify that the effectiveness of six image SSL baselines are all improved by learning hierarchical semantic information. Based on such comprehensive comparisons, we suggest {\method} as a general framework that can boost a wide range of image SSL approaches by learning hierarchical image semantics.

%%%%%%%%%%%%%%%%%%%%%%%%%%%%%%%%%%%%%%%%%%%%%%%%%%%%%%%%%%%%
%%%%%%%%%%%%%%%%%%%%%%%%%%%%%%%%%%%%%%%%%%%%%%%%%%%%%%%%%%%%

\vspace{-1.1mm}
\section{Related Work} \label{sec:rela}
\vspace{-1.1mm}

\textbf{CNN based self-supervised image representation learning.} Self-supervised learning (SSL) methods have been broadly studied to acquire effective image representations. Most of the prevailing approaches in this area are designed upon CNN architectures (\emph{e.g.}, ResNet~\cite{resnet}). In this line of research, most early works~\cite{colorize_1,colorize_2,rotation,jigsaw} focus on designing useful pretext tasks.
% \emph{e.g.}, colorizing gray-scale images~\cite{colorize_1,colorize_2}, recovering rotated images~\cite{rotation}, solving jigsaw puzzles~\cite{jigsaw}, \emph{etc.} 
Breakthroughs have been witnessed in this area since the introduction of \emph{contrastive learning}~\cite{cpc,npid,moco,simclr,moco_v2}. The key idea of contrastive learning is to embed similar instances nearby in the latent space while embed dissimilar ones distantly. Such an objective is commonly pursued by 
% maximizing the mutual information between different views of the same image, \emph{i.e.}, 
minimizing the InfoNCE loss~\cite{cpc,deepinfomax}. Recent works seek to improve the standard contrastive learning scheme by relieving the reliance on negative sampling~\cite{byol,simsiam,barlowtwins}, capturing the semantic information of image clusters~\cite{swav,pcl,hcsc}, utilizing stronger augmentation  functions~\cite{swav,clsa} or selecting better positive and negative pairs~\cite{debias,adco,infd}. These efforts enable the CNN model learned with self-supervision to have competitive performance with its fully-supervised counterpart. 
\vspace{-0.6mm}

\textbf{Vision Transformer based self-supervised image representation learning.} The development of different Vision Transformers~\cite{vit,deit,swin_transformer} revolutionize the basic architecture of image encoders. These architectures 
% are equipped with much less inductive bias than conventional CNN architectures and thus 
require specific schemes for self-supervised learning. Inspired by the masked language modeling (MLM) approach~\cite{bert} for pre-training text Transformers, masked image modeling (MIM) based methods~\cite{igpt,beit,mae,simmim,ibot} are proposed to learn self-supervised image representations with Vision Transformers. Besides MIM, self-distillation~\cite{dino,ibot,data2vec,context_ae} and contrastive learning~\cite{moco_v3} based objectives are also explored for the SSL of Vision Transformers. 
The performance of these approaches is catching up with fully-supervised learning. 
% These approaches are demonstrated to be effective under different Vision Transformer architectures, whose performance approaches fully-supervised learning. 
\vspace{-0.6mm}

\textbf{Hierarchical image representation learning.} Our work is closely related to the previous works~\cite{xie2016unsupervised,yang2016joint,deepcluster,deepercluster,sela,swav,pcl,hcsc} for simultaneous image clustering and representation learning. Most of these works~\cite{xie2016unsupervised,yang2016joint,deepcluster,sela,swav} learn a single level of semantics for an image dataset, which cannot capture the hierarchical semantic information that naturally exists underlying the images. Some recent works~\cite{deepercluster,pcl,hcsc} make an attempt on learning such hierarchical semantics, and they have achieved performance gain by injecting into specific SSL baselines. However, there still lacks a systematic study on whether hierarchical image representation learning can enhance various existing image SSL methods under different model architectures. 

To answer this question, this work proposes a general framework for hierarchical image representation learning, named as {\method}. Inspired by a probabilistic factorization, this framework naturally embraces the objective of an arbitrary off-the-shelf image SSL method and the objective for learning hierarchical semantic information. Under the {\method} framework, performance gain is observed on six representative image SSL methods (three CNN based and three Vision Transformer based), which, for the first time, verifies the general effectiveness of learning multiple levels of image semantics. 

%%%%%%%%%%%%%%%%%%%%%%%%%%%%%%%%%%%%%%%%%%%%%%%%%%%%%%%%%%%%
%%%%%%%%%%%%%%%%%%%%%%%%%%%%%%%%%%%%%%%%%%%%%%%%%%%%%%%%%%%%

\vspace{-0.5mm}
\section{Problem Definition and Preliminary} \label{sec:pre}
\vspace{-0.8mm}

\subsection{Problem Definition} \label{sec:pre_1}

Upon a set $X = \{x_1, x_2, \cdots, x_N\}$ of $N$ raw images without label information, we seek to learn a \emph{hierarchical representation} $z_n$ for each image $x_n \in X$. Specifically, we consider $L+1$ latent spaces $\{V_l\}_{l=0}^L$ that represent different levels of semantic information, and these latent semantics are desired to be organized in a \emph{fine-to-coarse} way, \emph{i.e.}, representing the most fine-grained semantics in the space $V_0$ and representing gradually more coarse-grained semantics in the remaining $L$ spaces. In this way, we represent each image with a chain $z_n = \{z_n^l\}_{l=0}^L$ of low-dimensional vectors, where $z_n^l \in \mathbb{R}^{d}$ denotes the representation of image $x_n$ in the semantic space $V_l$. Following the philosophy of self-supervised learning, the hierarchical image representations $Z = \{z_1, z_2, \cdots, z_N\}$ are learned under the supervision of the data itself. 

%%%%%%%%%%%%%%%%%%%%%%%%%%%%%%%%%%%%%%%%%%%%%%%%%%%%%%%%%%%%

\begin{wrapfigure}{R}{0.58\textwidth}
\begin{minipage}{0.58\textwidth}
  \vspace{-15.5mm}
  \begin{algorithm}[H]
    \caption{Hierarchical K-means.} \label{algo:hkmeans}
    \begin{spacing}{1}
    \begin{algorithmic}
      \STATE {\bfseries Input:} Vanilla image representations $\tilde{Z}$.
      \STATE {\bfseries Output:} Hierarchical prototypes $\mathcal{C}=\{\{c_i^l\}_{i=1}^{M_l}\}_{l=1}^L$, the set $\mathcal{E}$ of edges between different prototypes.
      \STATE $\{c_i^1\}_{i=1}^{M_1} \gets \textrm{K-means}(\tilde{Z})$.
      \FOR{$l=2$ {\bfseries to} $L$}
      \STATE $\{ c^l_i \}_{i=1}^{M_l} \gets \textrm{K-means}\big( \{ c^{l-1}_i \}_{i=1}^{M_{l-1}} \big)$.
      \FOR{$i=1$ {\bfseries to} $M_{l-1}$}
      \STATE $\mathcal{E} \gets \mathcal{E} \cup \big\{ \big( c^{l-1}_i, \textrm{AssignedPrototype}(c^{l-1}_i) \big) \big\}$.
      \ENDFOR
      \ENDFOR
    \end{algorithmic}
    \end{spacing}
  \end{algorithm}
  \vspace{-7mm}
\end{minipage}
\end{wrapfigure}

%%%%%%%%%%%%%%%%%%%%%%%%%%%%%%%%%%%%%%%%%%%%%%%%%%%%%%%%%%%%

\vspace{-0.5mm}
\subsection{Preliminary} \label{sec:pre_2}
\vspace{-0.5mm}

\begin{figure}[t]
\centering
    \includegraphics[width=1.0\linewidth]{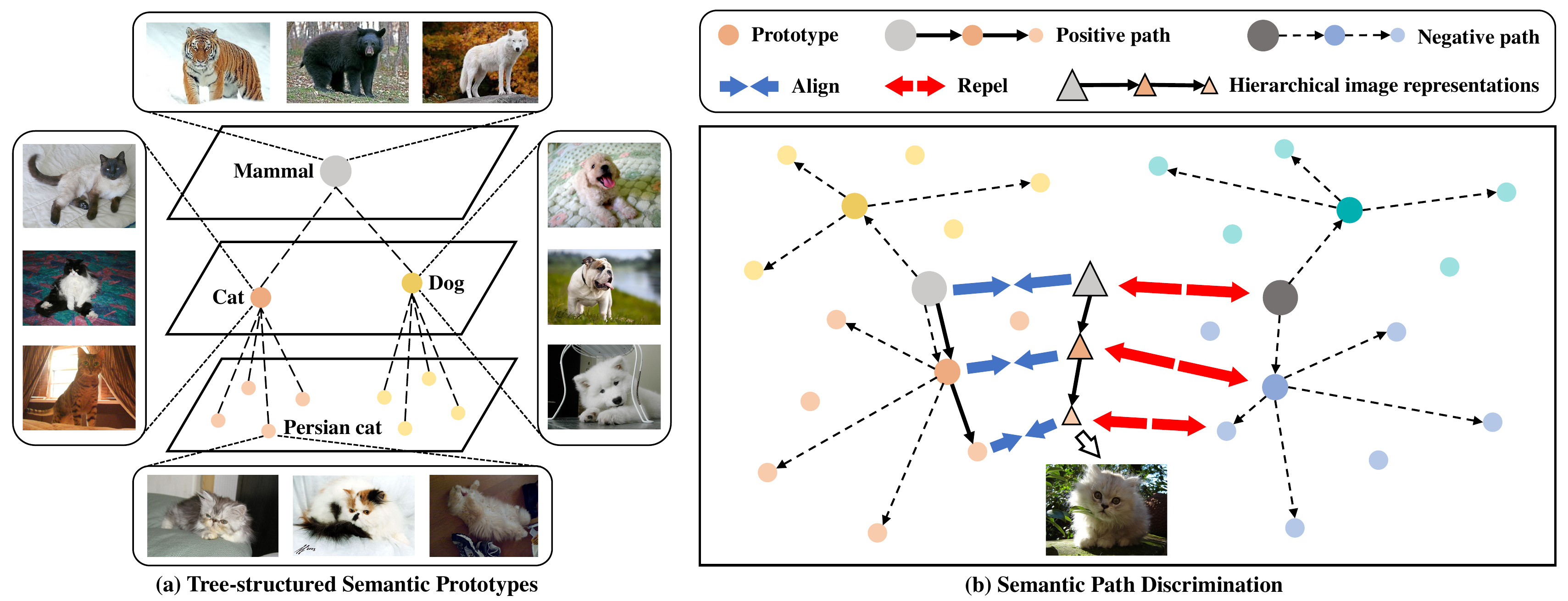}
    \vspace{-5.6mm}
    \caption{\textbf{Illustration of {\method} framework}. (a) Tree-structured semantic prototypes are constructed to guide the learning of hierarchical semantics of images. (b) Semantic path discrimination learns hierarchical image representations by aligning the representations with the corresponding positive semantic path while repelling the representations from negative semantic paths.}
    \label{fig:framework}
    \vspace{-1.8mm}
\end{figure}

\textbf{Hierarchical prototypes.} The learning of hierarchical image representations is guided by hierarchical prototypes~\cite{hcsc}, \emph{i.e.}, the representative embeddings of semantic clusters on different semantic levels, denoted as $\mathcal{C}=\{\{c_i^l\}_{i=1}^{M_l}\}_{l=1}^L$ ($L$: number of semantic levels; $M_l$: number of prototypes at the $l$-th semantic level). These prototypes are obtained by the hierarchical K-means algorithm~\cite{hcsc}, as summarized in Alg.~\ref{algo:hkmeans}. To start this procedure, we extract the vanilla image representations without hierarchical semantics by an image encoder, \emph{e.g.}, the average pooled embedding from ResNet~\cite{resnet} or the \texttt{[CLS]} token embedding from ViT~\cite{vit}, denoted as $\tilde{Z}=\{\tilde{z}_1, \tilde{z}_2, \cdots, \tilde{z}_N\}$. A standard K-means clustering is applied upon these image representations to derive the prototypes at the first semantic level. After that, the K-means clustering is iteratively applied to the prototypes at the current semantic level to get the prototypes at a more coarse-grained semantic level. The cluster assignment relations between two consecutive levels of prototypes are stored as an edge set $\mathcal{E}$, which structures hierarchical prototypes as a set of trees (Fig.~\ref{fig:framework}(a) shows such a tree). The hierarchical prototypes derived in this way form a fine-to-coarse semantic hierarchy, which is suitable to guide the learning of hierarchical image representations.

%%%%%%%%%%%%%%%%%%%%%%%%%%%%%%%%%%%%%%%%%%%%%%%%%%%%%%%%%%%%
%%%%%%%%%%%%%%%%%%%%%%%%%%%%%%%%%%%%%%%%%%%%%%%%%%%%%%%%%%%%

\vspace{-0.5mm}
\section{Hierarchical Image Representation Learning} \label{sec:method}
\vspace{-0.3mm}

In this section, we introduce the framework of hierarchical image representation learning ({\method}). This framework models hierarchical image representations in a factorized way, \emph{i.e.}, the modeling at the most fine-grained semantic level and the modeling at more coarse-grained semantic levels (Sec.~\ref{sec:method_1}). The first modeling problem can be solved by any off-the-shelf image SSL method, and the second modeling problem is solved by a novel \emph{semantic path discrimination} approach (Sec.~\ref{sec:method_2}). By combining these two kinds of objectives, HIRL can enhance existing image SSL methods through better capturing hierarchical semantic information (Sec.~\ref{sec:method_3}).  

%%%%%%%%%%%%%%%%%%%%%%%%%%%%%%%%%%%%%%%%%%%%%%%%%%%%%%%%%%%%

%%%%%%%%%%%%%%%%%%%%%%%%%%%%%%%%%%%%%%%%%%%%%%%%%%%%%%%%%%%%

\vspace{-0.5mm}
\subsection{Probabilistic Formalization} \label{sec:method_1}
\vspace{-0.5mm}

Given an image $x$, we aim to model its hierarchical representations $z=\{z^l\}_{l=0}^{L}$ that encode its semantic information in a fine-to-coarse way. Directly modeling the joint distribution $p(z^0, \cdots, z^L | x)$ of all semantic representations is hard, since there lacks a concept of the most fine-grained semantics that we should start from. Therefore, the {\method} framework solves this problem in a factorized way, \emph{i.e.}, first modeling $p(z^0|x)$ to set up the most fine-grained semantic representation and then modeling the joint distribution $p(z^1, \cdots, z^L | z^0)$ for more coarse-grained semantics:
\begin{equation} \label{eq1}
p(z^0, \cdots, z^L | x) = p(z^0 | x) p(z^1, \cdots, z^L | z^0) .
\end{equation}

The first term $p(z^0|x)$ can be suitably modeled by off-the-shelf image SSL methods, since they are designed to capture a single level of semantic information, \emph{e.g.}, by modeling the dependency among image patches~\cite{mae,ibot}, capturing the semantic similarity between image pairs~\cite{moco_v2,moco_v3} or modeling image clustering patterns~\cite{swav,dino}. However, the modeling of the second term $p(z^1, \cdots, z^L | z^0)$ for hierarchical semantic representations is non-trivial, which requires to capture the dependency of coarse-grained semantics on fine-grained semantics and also the interdependency among different levels of coarse-grained semantics. We introduce our solution in the next part. 

%%%%%%%%%%%%%%%%%%%%%%%%%%%%%%%%%%%%%%%%%%%%%%%%%%%%%%%%%%%%

\vspace{-0.5mm}
\subsection{Semantic Path Discrimination} \label{sec:method_2}
\vspace{-0.5mm}

In {\method}, we regard the image representation learned by an off-the-shelf SSL method as the most fine-grained semantic representation $z^0$ that lies in the space $V_0$. On such basis, we further learn hierarchical semantic representations $\{z^l\}_{l=1}^L$ to improve model's representation learning ability.

\textbf{Hierarchical representation derivation.} Upon the representation $z^0$, we utilize an MLP projection head $h_l: V_0 \rightarrow V_l$ to map it onto the semantic space $V_l$ representing more coarse-grained semantics,
\emph{i.e.}, $z^l = h_l (z^0)$ ($1 \leqslant l \leqslant L$). Learning these hierarchical representations requires some guidance on what specific semantic information should be encoded in each representation. We resort to \emph{hierarchical prototypes} (Sec.~\ref{sec:pre_2}) for such guidance. These prototypes serve as the anchor points that locate potential semantic clusters on different semantic spaces. In the derivation of hierarchical prototypes (Alg.~\ref{algo:hkmeans}), we set $\tilde{Z} = \{z^0_1, z^0_2, \cdots, z^0_N\}$ to guarantee that the modeling of coarse-grained semantics is based on the most fine-grained one, \emph{i.e.}, modeling $p(z^1, \cdots, z^L | z^0)$ faithfully. 

The other important part of modeling $p(z^1, \cdots, z^L | z^0)$ is to model the \emph{joint likelihood} of hierarchical semantic representations $\{z^l\}_{l=1}^L$ for a specific sample. To perform such joint modeling, we employ the consecutive prototypes (\emph{i.e.}, paths) in hierarchical prototypes as references and seek to align the representations $\{z^l\}_{l=1}^L$ as a whole with the most likely path for each sample. 

\textbf{Semantic path retrieval.} In hierarchical prototypes, we define a \emph{semantic path} $\mathbb{P}=\{c^l\}_{l=1}^L$ as a path traversing from a bottom layer prototype $c^1$ to its corresponding top layer prototype $c^L$ based on the tree structure. Such a path represents some hierarchical semantics underlying the data, like Persian cat $\rightarrow$ cat $\rightarrow$ mammal, which is desired to be encoded in hierarchical representations $\{z^l\}_{l=1}^L$. 

Given an image $x$, we define two kinds of paths in hierarchical prototypes, \emph{i.e.}, the \emph{positive path} and the \emph{negative path}, to learn its representations. The prototypes on the positive path encode the same semantics as $x$ on each semantic level, while the prototypes on the negative path could encode different semantics with $x$ on some semantic level. We extract the positive path $\mathbb{P}_+=\{c_+^l\}_{l=1}^L$ by first retrieving the bottom layer prototype $c_+^1$ that the image is assigned to during K-means clustering (the first step in Alg.~\ref{algo:hkmeans}) and then traversing from bottom to top. A negative path $\mathbb{P}_-=\{c_-^l\}_{l=1}^L$ starts from any bottom layer prototype $c_-^1$ other than $c_+^1$ and traverses from bottom to top. Note that, there exists a single positive path and $M_1 - 1$ possible negative paths for each image. 

\textbf{Semantic path discrimination.} Based on these definitions, we formalize the semantic path discrimination problem as maximizing the similarity between hierarchical representations $\{z^l\}_{l=1}^L$ and the positive path, while minimizing the similarity between these representations and the negative path. To measure the similarity between hierarchical representations $\{z^l\}_{l=1}^L$ and a semantic path $\mathbb{P}=\{c^l\}_{l=1}^L$, we compute the similarity between each corresponding pair $(z^l, c^l)$ of representation and prototype and multiplying across all pairwise similarities:
\begin{equation} \label{eq2}
\vspace{0.1mm}
s\big( \{z^l\}_{l=1}^L, \mathbb{P} \big) = \prod_{l=1}^L \frac{1 + \cos(z^l, c^l)}{2} ,
\vspace{0.1mm}
\end{equation}
where we map each cosine similarity $\cos(z^l, c^l)$ to $[0,1]$ for similarity multiplication. We randomly sample $N_{\mathrm{neg}}$ negative paths from all possible negative paths, and the objective function of semantic path discrimination is formalized as a balanced binary classification loss:
\begin{equation} \label{eq3}
\vspace{0.1mm}
\mathcal{L}_{\mathrm{SPD}} (\theta) = - \mathbb{E}_{x \sim p_d} \Big[ \log \Big( s\big( \{z^l\}_{l=1}^L, \mathbb{P}_+ \big) \Big) + \frac{1}{N_{\mathrm{neg}}} \sum_{k=1}^{N_{\mathrm{neg}}} \log \Big( 1 - s\big( \{z^l\}_{l=1}^L, \mathbb{P}^k_- \big) \Big) \Big] ,
\vspace{0.1mm}
\end{equation}
where $\theta$ denotes all model parameters including the parameters of image encoder and projection heads, and $p_d$ is the data distribution that all images are drawn from. We provide a graphical illustration of semantic path discrimination in Fig.~\ref{fig:framework}.

\textbf{Overall objective.} Under the {\method} framework, the joint distribution $p(z^0, \cdots, z^L | x)$ of all semantic representations is modeled in a factorized way. The objective function $\mathcal{L}_{\mathrm{img}}$ of an off-the-shelf image SSL method models the distribution $p(z^0|x)$ for the most fine-grained semantics, and the objective function $\mathcal{L}_{\mathrm{SPD}}$ of semantic path discrimination models the distribution $p(z^1, \cdots, z^L | z^0)$ for more coarse-grained semantics. The overall objective combines these two objectives to learn the hierarchical image representations on all considered semantic levels:
\begin{equation} \label{eq4}
\vspace{0.1mm}
\min \limits_{\theta} \, \mathcal{L}_{\mathrm{img}} (\theta) + \mathcal{L}_{\mathrm{SPD}} (\theta) .
\end{equation}

%%%%%%%%%%%%%%%%%%%%%%%%%%%%%%%%%%%%%%%%%%%%%%%%%%%%%%%%%%%%

\begin{wrapfigure}{R}{0.58\textwidth}
\begin{minipage}{0.58\textwidth}
  \vspace{-8.8mm}
  \begin{algorithm}[H]
    \caption{The learning procedure of HIRL.} \label{algo:learning}
    \begin{spacing}{1}
    \begin{algorithmic}
      \STATE {\bfseries Input:} Unlabeled images $X$, first-stage training epochs $T_1$, second-stage training epochs $T_2$.
      \STATE {\bfseries Output:} Pre-trained model $F_{\theta}$.
      \STATE Train the model with only $\mathcal{L}_{\mathrm{img}}$ for $T_1$ epochs.
      \FOR{$t=1$ {\bfseries to} $T_2$}
      \STATE $\tilde{Z} \gets F_{\theta} (X)$.
      \STATE $\mathcal{C} \gets \textrm{Hierarchical\_K-means}(\tilde{Z})$.
      \STATE Train the model with $\mathcal{L}_{\mathrm{img}}$ and $\mathcal{L}_{\mathrm{SPD}}$ for one epoch.
      \ENDFOR
    \end{algorithmic}
    \end{spacing}
  \end{algorithm}
  \vspace{-7mm}
\end{minipage}
\end{wrapfigure}

%%%%%%%%%%%%%%%%%%%%%%%%%%%%%%%%%%%%%%%%%%%%%%%%%%%%%%%%%%%%

\vspace{-1.6mm}
\subsection{Pre-training and Downstream Application} \label{sec:method_3}
\vspace{-0.5mm}

\textbf{Pre-training.} The pre-training under {\method} is performed by two stages, as summarized in Alg.~\ref{algo:learning}. In the first stage, the image encoder is trained with only $\mathcal{L}_{\mathrm{img}}$ to learn the most fine-grained semantic representation $z^0$, which sets up the basic concepts about image semantics. At the start of each epoch in the second stage, a hierarchical K-means step is performed upon the image representations $\tilde{Z} = \{z^0_1, z^0_2, \cdots, z^0_N\}$ extracted by the current encoder to update hierarchical prototypes, which balances between training efficiency and the accuracy of hierarchical prototypes as representative cluster embeddings. Its time complexity is analyzed in Sec.~\ref{sec:analysis_2}. In the second stage, $\mathcal{L}_{\mathrm{img}}$ and $\mathcal{L}_{\mathrm{SPD}}$ are jointly applied to train the encoder and projection heads, which facilitates the model to capture the hierarchical semantics of images. By comparison, existing image SSL methods mainly focus on capturing a single level of semantic information and are thus less effective. 

\textbf{Downstream application.} After pre-training under the {\method} framework, we can obtain a strong image encoder whose output representation $z^0$ can (1) encode fine-grained semantics by the modeling of $p(z^0|x)$ and also (2) implicitly encode different levels of coarse-grained semantic information by the modeling of $p(z^1, \cdots, z^L | z^0)$. Because of such merits, in downstream applications, we can discard the projection heads used in pre-training and only use the image encoder for feature extraction, which guarantees the transportability of the pre-trained model on various downstream tasks. 

%%%%%%%%%%%%%%%%%%%%%%%%%%%%%%%%%%%%%%%%%%%%%%%%%%%%%%%%%%%%
%%%%%%%%%%%%%%%%%%%%%%%%%%%%%%%%%%%%%%%%%%%%%%%%%%%%%%%%%%%%

\section{Experiments} \label{sec:exp}

\subsection{Implementation Details} \label{sec:exp_1}

\subsubsection{Image SSL Baselines} \label{sec:exp_1_1}

We adopt three representative CNN based image SSL methods (\emph{i.e.}, MoCo v2~\cite{moco_v2}, SimSiam~\cite{simsiam} and SwAV~\cite{swav}) and three representative Vision Transformer based image SSL methods (\emph{i.e.}, MoCo v3~\cite{moco_v3}, DINO~\cite{dino} and iBOT~\cite{ibot}) as baselines. To facilitate fair comparison and reproducibility, under PyTorch~\cite{pytorch} (BSD License), we re-implement these baselines and associated downstream tasks in a unified codebase (\url{https://github.com/hirl-team/HIRL}). We briefly introduce these baselines from the perspective of learning image semantics as below:
\begin{itemize}
    \item \textbf{MoCo v2~\cite{moco_v2}} (\emph{CNN based}) is a typical approach to acquire the semantic similarity of image pairs. It employs a standard contrastive learning scheme to maximize the similarity of correlated image pairs while minimize the similarity of uncorrelated image pairs. 
    \item \textbf{SimSiam~\cite{simsiam}} (\emph{CNN based}) is another popular method that learns pairwise semantic similarity. Instead of contrasting between positive and negative pairs as in MoCo v2, SimSiam directly maximizes the similarity between correlated (\emph{i.e.}, positive) image pairs, using no negative pairs. It introduces a stop-gradient scheme to avoid collapsed representations. 
    \item \textbf{SwAV~\cite{swav}} (\emph{CNN based}) is a superior image SSL approach for CNNs, which learns the semantic information encoded in a single hierarchy of image clusters. It designs a swapped prediction problem to predict the cluster assignment of a view of an image from its another view. It proposes a multi-crop augmentation scheme to further enhance model's performance. 
    \item \textbf{MoCo v3~\cite{moco_v3}} (\emph{Vision Transformer based}) can be regarded as a variant of MoCo v2 under Vision Transformer architectures, and it also learns pairwise semantic similarity. To stabilize the training process of SSL, it proposes a bunch of implementation suggestions that work well in practice. 
    \item \textbf{DINO~\cite{dino}} (\emph{Vision Transformer based}) learns image representations by self-distillation, \emph{i.e.}, aligning the categorical distribution output by a student network with that of a teacher network. In this way, DINO captures the semantics underlying a single hierarchy of image clusters. 
    \item \textbf{iBOT~\cite{ibot}} (\emph{Vision Transformer based}) employs a masked image modeling objective and a self-distillation objective. The masked image modeling objective seeks to align the categorical distribution of the same patch that is masked or unmasked in an image, and the self-distillation objective follows the one in DINO. By simultaneously optimizing these two objectives, iBOT can capture the semantic information encoded in a single hierarchy of image/patch clusters. 
\end{itemize}

%%%%%%%%%%%%%%%%%%%%%%%%%%%%%%%%%%%%%%%%%%%%%%%%%%%%%%%%%%%%

\subsubsection{Adaptation to the HIRL Framework} \label{sec:exp_1_2}

To study the effect of learning multiple levels of extra semantic information, we adapt each baseline method into the proposed {\method} framework and compare the performance before and after such adaptation. During adaptation, we follow the principle that all the model architectures, training procedures and hyperparameter settings of the baseline method are unchanged, such that fair comparison is ensured. The interaction between the baseline method and the {\method} framework mainly lies in two aspects, \emph{feature extraction} and \emph{loss computation}, which we briefly introduce here and leave more details in Sec.~\ref{supp_sec1_1}.

\textbf{Feature extraction.} Under the {\method} framework, we utilize the encoder of baseline to extract image representations $\tilde{Z}$ for the per-epoch hierarchical K-means step (Alg.~\ref{algo:learning}), and these representations also serve as the inputs of the projection heads $h_l$ ($1 \leqslant l \leqslant L$) for hierarchical semantic representations (Sec.~\ref{sec:method_2}). For MoCo v2, SimSiam and SwAV, each $\tilde{z} \in \tilde{Z}$ is the projected representation of the average pooled embedding from ResNet~\cite{resnet}. For MoCo v3, DINO and iBOT, each $\tilde{z} \in \tilde{Z}$ is the projected representation of the \texttt{[CLS]} token embedding from ViT~\cite{vit}. We employ the momentum encoder of MoCo v2 and the teacher network of DINO and iBOT for feature extraction, since they are used for downstream evaluation in official implementations.

\textbf{Loss computation.} In the second training stage of {\method} (Alg.~\ref{algo:learning}), we optimize the image encoder and the projection heads by a joint loss (Eq.~\eqref{eq4}) that combines the loss $\mathcal{L}_{\mathrm{img}}$ of baseline and the loss $\mathcal{L}_{\mathrm{SPD}}$ for learning hierarchical semantic representations. 

%%%%%%%%%%%%%%%%%%%%%%%%%%%%%%%%%%%%%%%%%%%%%%%%%%%%%%%%%%%%

\begin{wraptable}{r}{6.6cm}
\vspace{-15.3mm}
\caption{{\method} hyperparameters for baselines.} \label{tab:parameter}
\vspace{0.4mm}
\begin{adjustbox}{max width=1.0\linewidth}
\begin{tabular}{l|cc}
\toprule  
\bf{Method} & \bf{\# Prototypes} & \bf{\# Negative paths} \\
\midrule
\bf{MoCo v2} & (30000, 10000, 1000) & 1000 \\
\bf{SimSiam} & (30000, 10000, 1000) & 1000 \\
\bf{SwAV} & (30000, 10000, 1000) & 1000 \\
\bf{MoCo v3} & (30000, 10000, 5000) & 100 \\
\bf{DINO} & (30000, 10000, 5000) & 100 \\
\bf{iBOT} & (30000, 10000, 1000) & 100 \\
\bottomrule
\end{tabular}
\end{adjustbox}
\vspace{-4.5mm}
\end{wraptable} 

%%%%%%%%%%%%%%%%%%%%%%%%%%%%%%%%%

\subsubsection{Model Details} \label{sec:exp_1_3}

To fairly compare different CNN or Vision Transformer based SSL methods, we set ResNet-50~\cite{resnet} as the encoder for MoCo v2, SimSiam and SwAV and set ViT-B/16~\cite{vit} as the encoder for MoCo v3, DINO and iBOT. The projection heads $h_l$ ($1 \leqslant l \leqslant L$) are implemented as 2-layer MLPs with batch normalization and ReLU activation in between (for MoCo v3, 5-layer MLPs are used for better convergence). We use the same representation dimension $d$ for all semantic spaces $\{V_l\}_{l=0}^L$ (Tab.~\ref{tab:time} lists the dimension of each method). For each combination of a baseline method and {\method}, we set its number of semantic levels as $L=3$, and we search for its prototype numbers among $(M_1, M_2, M_3) \in \{(30000, 10000, 1000), (30000, 10000, 5000)\}$ and its negative path sampling size among $N_{\mathrm{neg}} \in \{100, 1000\}$. For this search, we hold out 8,000 ImageNet~\cite{imagenet} training samples for validation and select the configuration with the lowest joint loss (Eq.~\eqref{eq4}) on this validation split. The search results are shown in Tab.~\ref{tab:parameter}. We do ablation studies for $L$, $\{M_l\}_{l=1}^L$ and $N_{\mathrm{neg}}$ in Sec.~\ref{sec:analysis_1}. 

%%%%%%%%%%%%%%%%%%%%%%%%%%%%%%%%%%%%%%%%%%%%%%%%%%%%%%%%%%%%

\vspace{-0.7mm}
\subsubsection{Training Details} \label{sec:exp_1_4}
\vspace{-0.5mm}

For both a baseline and its variant under the {\method} framework, we totally follow its official implementation to set the optimizer, learning rate and learning rate scheduler, and the detailed setups for each baseline are listed in Sec.~\ref{supp_sec1_2}. We train three CNN based approaches for 200 epochs and train three Vision Transformer based approaches for 400 epochs. The official SwAV suggests 800 epochs training for better performance, which we provide in Sec.~\ref{supp_sec_2}. The number of first-stage training epochs (\emph{i.e.}, $T_1$ in Alg.~\ref{algo:learning}) is set as 20 for MoCo v2, SimSiam, SwAV and MoCo v3 and is set as 40 for DINO and iBOT, and the second-stage training takes the rest epochs. We list the training batch size, GPU number and training time of each method in Tab.~\ref{tab:knn_linear_finetune}.

%%%%%%%%%%%%%%%%%%%%%%%%%%%%%%%%%%%%%%%%%%%%%%%%%%%%%%%%%%%%

\vspace{-0.9mm}
\subsection{Experimental Results\protect\footnote{For all tables in this section, we use `$\uparrow$', `$\rightarrow$' and `$\downarrow$' to denote improvement, equality and degeneration over baseline, respectively. \textbf{Bold results} are best within their model type, \emph{i.e.}, CNN or Vision Transformer.}} \label{sec:exp_2}

%%%%%%%%%%%%%%%%%%%%%%%%%%%%%%%%%%%%%%%%%%%%%%%%%%%%%%%%%%%%

\begin{table}[t]
\begin{spacing}{1.15}
    \centering
    \caption{Training environments and performance comparison on KNN evaluation (Top1-Acc), linear classification (Top1-Acc), fine-tuning (Top1-Acc) and semi-supervised learning with 1\% or 10\% data (Top1-Acc). We use Tesla-V100-32GB GPUs for all experiments.}
    \label{tab:knn_linear_finetune}
    \vspace{-1.8mm}
    \begin{adjustbox}{max width=1\linewidth}
        \begin{tabular}{l|ccccc|ccc|cc}
            \toprule
            \bf{Method} & \bf{Arch.} & \bf{\# Epochs} & \bf{Batch size} & \bf{\# GPUs} & \bf{Train time} & \bf{KNN} & \bf{Linear} & \bf{Fine-tune} & \bf{Semi-1\%} & \bf{Semi-10\%} \\
            \midrule \midrule
            \multicolumn{11}{c}{\bf{CNN based image SSL methods}} \\
            \midrule
            \bf{MoCo v2} & ResNet-50 & 200 & 256 & 8 & 45h 23min & 55.74 & 67.60 & 73.14 & 35.96 & 60.6 \\
            \bf{{\method}-MoCo v2} & ResNet-50 & 200 & 256 & 8 & 58h 50min & 57.56 ($\uparrow\!\;$1.82) & 68.40 ($\uparrow\!\;$0.8) & 73.86 ($\uparrow\!\;$0.72) & 37.31 ($\uparrow\!\;$1.35) & 61.71 ($\uparrow\!\;$1.11) \\
            \midrule
            \bf{SimSiam} & ResNet-50 & 200 & 512 & 16 & 40h 32min & 60.17 & 69.74 & 72.25 & 31.04 & 61.55 \\
            \bf{{\method}-SimSiam} & ResNet-50 & 200 & 512 & 16 & 70h 30min & 62.68 ($\uparrow\!\;$2.51) & 69.81 ($\uparrow\!\;$0.07) & 72.88 ($\uparrow\!\;$0.63) & 33.34 ($\uparrow\!\;$2.30) & 62.94 ($\uparrow\!\;$1.39) \\
            \midrule
            \bf{SwAV} & ResNet-50 & 200 & 4096 & 32 & 20h 15min & 63.45 & 72.68 & 76.82 & 51.23 & 68.72 \\
            \bf{{\method}-SwAV} & ResNet-50 & 200 & 4096 & 32 & 27h 45min & \textbf{63.99} ($\uparrow\!\;$0.54) & \textbf{73.43} ($\uparrow\!\;$0.75) & \textbf{77.18} ($\uparrow\!\;$0.36) & \textbf{52.12} ($\uparrow\!\;$0.89) & \textbf{69.32} ($\uparrow\!\;$0.60) \\
            \midrule
            \multicolumn{11}{c}{\bf{Vision Transformer based image SSL methods}} \\
            \midrule
            \bf{MoCo v3} & ViT-B/16 & 400 & 4096 & 128 & 26h 18min & 71.29 & 76.44 & 81.94 & 58.01 & 75.20 \\
            \bf{{\method}-MoCo v3} & ViT-B/16 & 400 & 4096 & 128 & 45h 20min & 71.68 ($\uparrow\!\;$0.39) & 75.12 ($\downarrow\!\;$1.32) & 82.19 ($\uparrow\!\;$0.25) & 60.92 ($\uparrow\!\;$2.91) & 74.91 ($\downarrow\!\;$0.29) \\
            \midrule
            \bf{DINO} & ViT-B/16 & 400 & 1024 & 32 & 86h 40min & 76.01 & 78.07 & 82.09 & 62.35 & 76.57 \\
            \bf{{\method}-DINO} & ViT-B/16 & 400 & 1024 & 32 & 107h 40min & 76.84 ($\uparrow\!\;$0.83) & 78.32 ($\uparrow\!\;$0.25) & 83.24 ($\uparrow\!\;$1.15) & 64.73 ($\uparrow\!\;$2.38) & 76.85 ($\uparrow\!\;$0.28) \\
            \midrule
            \bf{iBOT} & ViT-B/16 & 400 & 1024 & 32 & 87h 20min & 76.64 & 79.00 & 82.47 & 64.60 & 77.19 \\
            \bf{{\method}-iBOT} & ViT-B/16 & 400 & 1024 & 32 & 108h 20min & \textbf{77.49} ($\uparrow\!\;$0.85) & \textbf{79.36} ($\uparrow\!\;$0.36) & \textbf{83.37} ($\uparrow\!\;$0.90) & \textbf{66.15} ($\uparrow\!\;$1.55) & \textbf{77.32} ($\uparrow\!\;$0.13) \\
            \bottomrule
        \end{tabular}
    \end{adjustbox}
\end{spacing}
\vspace{-2.4mm}
\end{table}

%%%%%%%%%%%%%%%%%%%%%%%%%%%%%%%%%%%%%%%%%%%%%%%%%%%%%%%%%%%%

\vspace{-0.5mm}
\subsubsection{KNN Evaluation, Linear Classification and Fine-tuning} \label{sec:exp_2_1}
\vspace{-0.5mm}

\textbf{Evaluation Setups.} All the three tasks are evaluated on the ImageNet dataset~\cite{imagenet}. \emph{KNN evaluation} predicts each sample's label based on the labels of its nearest neighbors. We follow NPID~\cite{npid} to report the highest KNN classification accuracy among $K \in \{10, 20, 100, 200\}$. 

\emph{Linear classification} trains a single linear layer upon extracted image representations for classification. Since the parameter settings of linear classification vary among different baselines, we follow the official parameter settings of each baseline to train its variant under the {\method} framework. 

\emph{Fine-tuning} trains the image encoder along with a linear classifier for classification. We use two sets of parameters for CNN and Vision Transformer based methods, respectively. For CNN based methods, we train with an SGD optimizer (encoder learning rate: 0.1, classifier learning rate: 1.0, weight decay: 0, momentum: 0.9, batch size: 1024) for 20 epochs, and the learning rate is decayed by a factor of 0.2 at the 12th and 16th epoch. For Vision Transformer based methods, we train with an AdamW optimizer (initial learning rate: $4.0 \times 10^{-3}$, learning rate layer decay factor~\cite{beit}: 0.65, weight decay: 0.05, batch size: 1024) and a cosine annealing scheduler~\cite{cos} for 100 epochs.

% 1. knn: highest among K in {10,20,100,200}
% 2. linear: learn a single linear layer, mainly follow the baseline on learning, batch size, optimizer, learning rate scheduler
% 3. fine-tune: learn a single linear along with backbone; CNN: backbone lr 0.1, linear layer lr 1.0, batch size 1024, epochs 20, SGD (no wd, momentum 0.9), lr scheduler ([12, 16] decay 0.2); ViT: initial lr 0.004, layer decay 0.65, batch size 1024, epochs 100, AdamW (wd 0.05), lr scheduler cosine.

\textbf{Results.} In Tab.~\ref{tab:knn_linear_finetune}, we report the performance of six baselines and their variants under the {\method} framework. On all three tasks, SwAV and iBOT perform best within CNN and Vision Transformer based approaches, respectively. After adapting into the {\method} framework, performance gain is achieved on all methods and tasks except for {\method}-MoCo v3 on linear classification, and the superiority of SwAV and iBOT still preserves. These results demonstrate that {\method} can indeed boost various image SSL methods by learning extra hierarchical semantic information. 

%%%%%%%%%%%%%%%%%%%%%%%%%%%%%%%%%%%%%%%%%%%%%%%%%%%%%%%%%%%%

\vspace{-0.5mm}
\subsubsection{Semi-supervised Learning} \label{sec:exp_2_2}
\vspace{-0.5mm}

\textbf{Evaluation Setups.} \emph{Semi-supervised learning} trains the encoder and a linear classifier on 1\% or 10\% labeled data of ImageNet. Since MoCo v2 and SimSiam do not evaluate on this task originally, we adopt the official training configurations of SwAV for all three CNN based methods. The training configurations of three Vision Transformer based methods are adapted from their configurations for fine-tuning (Sec.~\ref{sec:exp_2_1}), in which only the initial learning rate is changed for better convergence (semi-1\% initial learning rate: $1.0 \times 10^{-3}$, semi-10\% initial learning rate: $1.0 \times 10^{-4}$). 

% 1. semi: CNN all use SwAV configs; ViT adapts config from fine-tune as only initial lr changed (1\%: 1e-3, 10\%: 1e-4).

\textbf{Results.} 
% Tab.~\ref{tab:knn_linear_finetune} presents the semi-supervised learning results of six baselines and their variants under {\method}. 
In Tab.~\ref{tab:knn_linear_finetune}, we observe that {\method} succeeds in enhancing the performance of all six baselines on both data settings except for {\method}-MoCo v3 on learning with 10\% data. In particular, iBOT is improved to achieve a new state-of-the-art performance. Therefore, by acquiring hierarchical image semantics, the pre-trained models show stronger ability of learning from insufficient data. 

%%%%%%%%%%%%%%%%%%%%%%%%%%%%%%%%%%%%%%%%%%%%%%%%%%%%%%%%%%%%

\begin{wraptable}{r}{7.8cm}
\vspace{-13mm}
\caption{Performance comparison on transfer learning.} \label{tab:transfer}
\vspace{0.4mm}
\begin{adjustbox}{max width=1.0\linewidth}
\begin{tabular}{l|ccc}
\toprule  
\bf{Method} & \bf{Places205 (Top1-Acc)} & \bf{COCO Det. (AP Box)} & \bf{COCO Seg. (AP Mask)} \\
\midrule \midrule
\multicolumn{4}{c}{\bf{CNN based image SSL methods}} \\
\midrule
\bf{MoCo v2} & 61.4 & 40.6 & 35.5 \\
\bf{{\method}-MoCo v2} & 61.9 ($\uparrow\!\;$0.5) & \textbf{41.1} ($\uparrow\!\;$0.5) & \textbf{35.7} ($\uparrow\!\;$0.2) \\
\midrule
\bf{SimSiam} & 58.4 & 37.6 & 33.1 \\
\bf{{\method}-SimSiam} & 61.3 ($\uparrow\!\;$2.9) & 40.0 ($\uparrow\!\;$2.4) & 35.0 ($\uparrow\!\;$1.9) \\
\midrule
\bf{SwAV} & 62.8 & 39.8 & 34.6 \\
\bf{{\method}-SwAV} & \textbf{63.2} ($\uparrow\!\;$0.4) & 40.5 ($\uparrow\!\;$0.7) & 35.2 ($\uparrow\!\;$0.6) \\
\midrule
\multicolumn{4}{c}{\bf{Vision Transformer based image SSL methods}} \\
\midrule
\bf{MoCo v3} & 66.5 & 48.0 & 41.7 \\
\bf{{\method}-MoCo v3} & 66.6 ($\uparrow\!\;$0.1) & 48.4 ($\uparrow\!\;$0.4) & 42.0 ($\uparrow\!\;$0.3) \\
\midrule
\bf{DINO} & 66.6 & 49.7 & 43.1 \\
\bf{{\method}-DINO} & 67.1 ($\uparrow\!\;$0.5) & 50.3 ($\uparrow\!\;$0.6) & 43.7 ($\uparrow\!\;$0.6) \\
\midrule
\bf{iBOT} & 67.3 & 50.9 & 43.9 \\
\bf{{\method}-iBOT} & \textbf{67.7} ($\uparrow\!\;$0.4) & \textbf{51.1} ($\uparrow\!\;$0.2) & \textbf{44.0} ($\uparrow\!\;$0.1) \\
\bottomrule
\end{tabular}
\end{adjustbox}
\vspace{-4.5mm}
\end{wraptable} 

%%%%%%%%%%%%%%%%%%%%%%%%%%%%%%%%%%%%%%%%%%%%%%%%%%%%%%%%%%%%

\vspace{-0.6mm}
\subsubsection{Transfer Learning} \label{sec:exp_2_3}
\vspace{-0.5mm}

\textbf{Evaluation Setups.} We evaluate on three transfer learning tasks, \emph{i.e.}, classification on Places205~\cite{places205}, object detection on COCO~\cite{coco} and instance segmentation on COCO~\cite{coco}. The configurations of the classification task follow those of fine-tuning (Sec.~\ref{sec:exp_2_1}). For detection and segmentation, the official configurations of MoCo v2 are applied to all CNN based methods, and the official configurations of iBOT are applied to all Vision Transformer based methods. More evaluation details can be found in the Sec.~\ref{supp_sec1_3}. 

% 1. places: all methods follow finetune.
% 2. Detection: all CNN follow MoCo v2 config; ViT all follow iBOT.
% 3. Segmentation: all CNN follow MoCo v2 config; ViT all follow iBOT.

\textbf{Results.} According to the comparisons in Tab.~\ref{tab:transfer}, all baselines achieve performance gain on all three transfer learning tasks after adapted into {\method}. These improvements verify the strength of {\method} on learning transferable image representations. 

%%%%%%%%%%%%%%%%%%%%%%%%%%%%%%%%%%%%%%%%%%%%%%%%%%%%%%%%%%%%

\begin{wraptable}{r}{5.2cm}
\vspace{-13.5mm}
\caption{Performance comparison on clustering evaluation.} \label{tab:cluster}
\vspace{0.4mm}
\begin{adjustbox}{max width={1.0\linewidth}, max height={0.1\textheight}}
\begin{tabular}{l|ccc}
\toprule  
\bf{Method} & \bf{Acc} & \bf{NMI} & \bf{AMI} \\
\midrule \midrule
\multicolumn{4}{c}{\bf{CNN based image SSL methods}} \\
\midrule
\bf{MoCo v2} & 29.3 & 66.6 & 41.5 \\
\bf{{\method}-MoCo v2} & 32.9 ($\uparrow\!\;$3.6) & \textbf{69.4} ($\uparrow\!\;$2.8) & \textbf{46.5} ($\uparrow\!\;$5.0)\\
\midrule
\bf{SimSiam} & 27.9 & 65.4 & 39.2 \\
\bf{{\method}-SimSiam} & 30.2 ($\uparrow\!\;$2.3) & 66.7 ($\uparrow\!\;$1.3) & 41.7 ($\uparrow\!\;$2.5) \\
\midrule
\bf{SwAV} & 32.0 & 67.4 & 43.0 \\
\bf{{\method}-SwAV} & \textbf{33.5} ($\uparrow\!\;$1.5) & 68.7 ($\uparrow\!\;$1.3) & 45.3 ($\uparrow\!\;$2.3) \\
\midrule
\multicolumn{4}{c}{\bf{Vision Transformer based image SSL methods}} \\
\midrule
\bf{MoCo v3} & 50.0 & 78.2 & 61.9 \\
\bf{{\method}-MoCo v3} & 50.2 ($\uparrow\!\;$0.2) & 79.4 ($\uparrow\!\;$1.2) & 64.4 ($\uparrow\!\;$2.5) \\
\midrule
\bf{DINO} & 50.2 & 80.8 & 67.4 \\   
\bf{{\method}-DINO} & 52.4 ($\uparrow\!\;$2.2) & 80.8 ($\rightarrow$) & 66.6 ($\downarrow\!\;$0.8) \\
\midrule
\bf{iBOT} & 51.1 & 79.9  & 64.9 \\
\bf{{\method}-iBOT} & \textbf{53.4} ($\uparrow\!\;$2.3) & \textbf{81.6} ($\uparrow\!\;$1.7) & \textbf{68.0} ($\uparrow\!\;$3.1) \\
\bottomrule
\end{tabular}
\end{adjustbox}
\vspace{-4.5mm}
\end{wraptable} 

%%%%%%%%%%%%%%%%%%%%%%%%%%%%%%%%%%%%%%%%%%%%%%%%%%%%%%%%%%%%

\vspace{-0.4mm}
\subsubsection{Clustering Evaluation} \label{sec:exp_2_4}
\vspace{-0.5mm}

\textbf{Evaluation Setups.} Following iBOT~\cite{ibot}, we conduct clustering evaluation on the validation set of ImageNet with 1,000 clusters and measure with three metrics, \emph{i.e.}, accuracy, normalized mutual information (NMI) and adjusted mutual information (AMI). For all CNN based methods, we perform K-means on the average pooled embedding from ResNet. For all Vision Transformer based methods, we perform K-means on the \texttt{[CLS]} token embedding from ViT. 

\textbf{Results.} As shown in Tab.~\ref{tab:cluster}, {\method} notably boosts the clustering performance of MoCo v2, SimSiam, SwAV, MoCo v3 and iBOT on all three metrics, and it enhances DINO on clustering accuracy. These improvements are mainly attributed to the proposed semantic path discrimination scheme, which can well capture different levels of semantic clusters underlying the raw images. 

%%%%%%%%%%%%%%%%%%%%%%%%%%%%%%%%%%%%%%%%%%%%%%%%%%%%%%%%%%%%
%%%%%%%%%%%%%%%%%%%%%%%%%%%%%%%%%%%%%%%%%%%%%%%%%%%%%%%%%%%%

\begin{figure}[t]
\centering
    \includegraphics[width=1.0\linewidth]{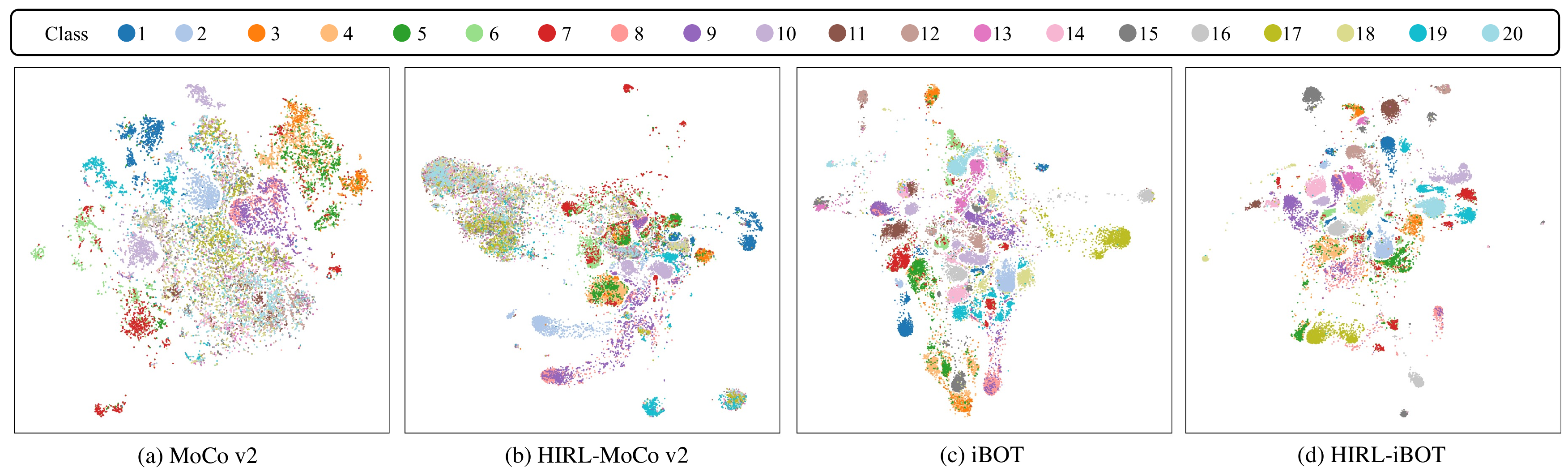}
    \vspace{-6.5mm}
    \caption{Feature visualization of ImageNet training samples from the first 20 classes.}
    \label{fig:visualization}
    \vspace{-1.8mm}
\end{figure}

%%%%%%%%%%%%%%%%%%%%%%%%%%%%%%%%%%%%%%%%%%%%%%%%%%%%%%%%%%%%

\section{Analysis} \label{sec:analysis}

%%%%%%%%%%%%%%%%%%%%%%%%%%%%%%%%%%%%%%%%%%%%%%%%%%%%%%%%%%%%

\begin{wraptable}{r}{6.0cm}
\vspace{-21.2mm}
\caption{Ablation study for prototype structure and negative path sampling. (All results are reported on {\method}-MoCo v2.)} \label{tab:ablation}
\vspace{0.4mm}
\begin{adjustbox}{max width={1.0\linewidth}, max height={0.1\textheight}}
\begin{tabular}{lc|c}
\toprule  
\bf{\# Prototypes} & \bf{\# Negative paths} & \bf{KNN (Top1-Acc)} \\
\midrule \midrule
(30000) & 1000 & 56.22 \\
(50000) & 1000 & 56.71 \\
(30000, 10000) & 1000 & 56.75 \\
(30000, 10000, 1000) & 1000 & 57.56 \\
(30000, 10000, 1000, 200) & 1000 & \textbf{57.72} \\
\midrule
(30000, 10000, 1000) & 100 & 57.28 \\
(30000, 10000, 1000) & 1000 & \textbf{57.56} \\
(30000, 10000, 1000) & 5000 & 57.37 \\
(30000, 10000, 1000) & 10000 & 57.32 \\
\bottomrule
\end{tabular}
\end{adjustbox}
\vspace{-3.8mm}
\end{wraptable} 

%%%%%%%%%%%%%%%%%%%%%%%%%%%%%%%%%%%%%%%%%%%%%%%%%%%%%%%%%%%%

\vspace{-1.5mm}
\subsection{Ablation Study} \label{sec:analysis_1}
\vspace{-0.8mm}

\textbf{Effect of prototype structure.} In the first block of Tab.~\ref{tab:ablation}, we evaluate the {\method}-MoCo v2 trained with different prototype structures. From the 1st, 3rd, 4th and 5th rows, we can observe a monotonic performance increase as the number of prototype layers increases, which demonstrates the effectiveness of learning more levels of semantic information. By comparing the 2nd and 5th rows, it is further shown that learning with multiple layers of prototypes is more effective than learning with a single layer of more prototypes (\emph{i.e.}, {\small $50000 > 30000 + 10000 + 1000 + 200$}).

\textbf{Effect of negative path sampling.} The second block of Tab.~\ref{tab:ablation} studies the effect of negative path sampling on {\method}-MoCo v2. The results illustrate that the objective function of semantic path discrimination (Eq.~\eqref{eq3}) is comparably effective under a wide range of negative path sampling sizes.

%%%%%%%%%%%%%%%%%%%%%%%%%%%%%%%%%%%%%%%%%%%%%%%%%%%%%%%%%%%%

\begin{wraptable}{r}{5.0cm}
\vspace{-14.7mm}
\caption{Running time comparison.} \label{tab:time}
\vspace{0.4mm}
\begin{adjustbox}{max width={1.0\linewidth}, max height={0.08\textheight}}
\begin{tabular}{lc|c}
\toprule  
\bf{Method} & $\quad\;$\bf{Dimension}$\quad\;$ & $\;\,$\bf{Per-epoch runtime}$\;\,$ \\
\midrule \midrule
\bf{MoCo v2} & 128 & 13min 39s \\
\bf{{\method}-MoCo v2} & 128 & 17min 30s \\
\midrule
\bf{SimSiam} & 2048 & 12min 9s \\
\bf{{\method}-SimSiam} & 2048 & 21min 10s \\
\midrule
\bf{SwAV} & 128 & 6min 3s \\
\bf{{\method}-SwAV} & 128 & 8min 15s \\
\midrule
\bf{MoCo v3} & 256 & 3min 58s \\
\bf{{\method}-MoCo v3} & 256 & 6min 30s \\
\midrule
\bf{DINO} & 256 & 13min 0s \\
\bf{{\method}-DINO} & 256 & 16min 9s \\
\midrule
\bf{iBOT} & 256 & 13min 6s \\
\bf{{\method}-iBOT} & 256 & 16min 10s \\
\bottomrule
\end{tabular}
\end{adjustbox}
\vspace{-5.0mm}
\end{wraptable} 

%%%%%%%%%%%%%%%%%%%%%%%%%%%%%%%%%%%%%%%%%%%%%%%%%%%%%%%%%%%%

\vspace{-0.8mm}
\subsection{Time Complexity Analysis} \label{sec:analysis_2}
\vspace{-0.5mm}

Compared to a baseline, its variant under {\method} requires an additional hierarchical K-means step per epoch, whose time complexity is $\mathcal{O}(N M_1 d + \sum_{l=1}^{L-1} M_l M_{l+1} d) = \mathcal{O}(N M_1 d)$ ($N$: dataset size, $M_l$: prototype number at the $l$-th semantic level, $d$: representation dimension). This step makes up of the majority of {\method}'s extra computation. Since $N$ and $M_1$ are identical among different methods, a larger $d$ will lead to higher extra cost, which is verified by the results in Tab.~\ref{tab:time}.
% The results in Tab.~\ref{tab:time} verify this point, where the extra per-epoch runtime of {\method}-SimSiam (dimension: 2048) is much longer than that of other methods (dimension: 128 or 256).

%%%%%%%%%%%%%%%%%%%%%%%%%%%%%%%%%%%%%%%%%%%%%%%%%%%%%%%%%%%%

\vspace{-1.3mm}
\subsection{Visualization} \label{sec:analysis_3}
\vspace{-0.8mm}

In Fig.~\ref{fig:visualization}, we use t-SNE~\cite{tsne} to visualize image representations. We randomly sample 2,000 images from each of the first 20 classes of ImageNet training set for visualization. Compared to the original MoCo v2 and iBOT, the image representations of different classes are better separated after trained under the {\method} framework. For example, the original iBOT confuses the representations of class 18 and 20, while these two classes are well divided by {\method}-iBOT in the latent space.

% In Fig.~\ref{fig:visualization}, we use t-SNE~\cite{tsne} to visualize the image representations extracted by MoCo v2, iBOT and their variants under {\method}. We randomly sample 2,000 images from each of the first 20 classes of ImageNet training set for visualization. Compared to the original MoCo v2 and iBOT, the image representations of different classes are better separated after trained under the {\method} framework. For example, the original iBOT confuses the representations of class 18 and 20, while these two classes are well divided by {\method}-iBOT in the latent space. These qualitative results demonstrate the effectiveness of {\method} on learning discriminative image representations. 

%%%%%%%%%%%%%%%%%%%%%%%%%%%%%%%%%%%%%%%%%%%%%%%%%%%%%%%%%%%%
%%%%%%%%%%%%%%%%%%%%%%%%%%%%%%%%%%%%%%%%%%%%%%%%%%%%%%%%%%%%

\vspace{-0.9mm}
\section{Conclusions and Future Work} \label{sec:conclusion}
\vspace{-0.9mm}

This work proposes a general framework for Hierarchical Image Representation Learning ({\method}). It acquires the most fine-grained semantics by an arbitrary off-the-shelf image SSL method, and it learns more coarse-grained semantic information by semantic path discrimination. {\method} succeeds in enhancing six representative image SSL approaches on a broad range of downstream tasks. 
% By combining with six representative image SSL approaches, {\method} succeeds in enhancing these baseline approaches on a broad range of downstream tasks. 

The main limitation of the current {\method} framework is that it can only learn the semantics of individual visual objects, while the interactions between different objects are not well captured. Therefore, our future work will mainly focus on designing an SSL method that can jointly capture the semantics of visual objects and their interactions. 
% The main limitation of the current {\method} framework is that it learns semantic representations from pure images. However, the data from other modalities, \emph{e.g.}, texts and audio, can always provide complementary semantic information about images. Therefore, our future work will mainly focus on utilizing the multi-modal data, like videos, to acquire more informative semantic representations. 

%%%%%%%%%%%%%%%%%%%%%%%%%%%%%%%%%%%%%%%%%%%%%%%%%%%%%%%%%%%%

%%%%%%%%%%%%%%%%%%%%%%%%%%%%%%%%%%%%%%%%%%%%%%%%%%%%%%%%%%%%

\newpage
\bibliographystyle{plain} 
\bibliography{reference.bib}

\newpage
\appendix

%%%%%%%%%%%%%%%%%%%%%%%%%%%%%%%%%%%%%%%%%%%%%%%%%%%%%%%%%%%%

\section{More Implementation Details} \label{supp_sec_1}
\vspace{-1.2mm}

%%%%%%%%%%%%%%%%%%%%%%%%%%%%%%%%%%%%%%%%%%%%%%%%%%%%%%%%%%%%

\subsection{Adaptation of Baselines to the HIRL Framework} \label{supp_sec1_1}
\vspace{-1.2mm}

% 1. hierarchical k-means applied on which embedding
% 2. swav: iterwise assignment (re-assign cluster id according to current image representation; inspired by skinhorn update)

In the main paper, we introduce how each baseline method is adapted into the {\method} framework from two aspects, feature extraction and loss computation. For feature extraction, {\method} deems the image representation extracted by the baseline model as the representation $z^0$ for most fine-grained semantics. Based on this representation, {\method} discovers hierarchical prototypes and derives the representations $\{z^l\}_{l=1}^L$ for more coarse-grained semantics. Here, we specifically state which layer's output of each baseline method serves as the representation $z^0$. 
\vspace{-0.3mm}
\begin{itemize}
\item \textbf{MoCo v2~\cite{moco_v2}} uses a 2-layer MLP to project the average pooled embedding of ResNet to the semantic space for contrastive learning. Under the {\method} framework, we regard the output of this projection head as $z^0$. 
\item \textbf{SimSiam~\cite{simsiam}} employs a 3-layer MLP to project the average pooled embedding of ResNet to the semantic space for cross-view prediction. We adopt this projection head's output as $z^0$. Upon the projection head, SimSiam involves another MLP as the predictor, while it is not used to derive hierarchical semantic representations.
\item \textbf{SwAV~\cite{swav}} uses a 2-layer MLP to project the average pooled embedding of ResNet to the semantic space for cluster assignment prediction. The projected embedding serves as $z^0$ under {\method}. 
\item \textbf{MoCo v3~\cite{moco_v3}} utilizes a 3-layer MLP to project the \texttt{[CLS]} token embedding of ViT to the semantic space for contrastive learning. We deem the output of this projection head as $z^0$. Similar as SimSiam, MoCo v3 defines an MLP predictor upon the projection head for cross-view prediction. This predictor is not used to derive hierarchical semantic representations.
\item \textbf{DINO~\cite{dino}} employs a 3-layer MLP to project the \texttt{[CLS]} token embedding of ViT to the semantic space for self-distillation. The output of this projection head is deemed as $z^0$ under {\method}.
\item \textbf{iBOT~\cite{ibot}} uses a 3-layer MLP to project the \texttt{[CLS]} token embedding of ViT to the semantic space for self-distillation and masked image modeling. We adopt the projected embedding as $z^0$.
\end{itemize}

%%%%%%%%%%%%%%%%%%%%%%%%%%%%%%%%%%%%%%%%%%%%%%%%%%%%%%%%%%%%

\vspace{-1.3mm}
\subsection{More Training Details} \label{supp_sec1_2}
\vspace{-2.3mm}

%%%%%%%%%%%%%%%%%%%%%%%%%%%%%%%%%%%%%%%%%%%%%%%%%%%%%%%%%%%%

\begin{table}[h]
\begin{spacing}{1.15}
    \centering
    \caption{More training configurations of baselines and their variants under {\method}.}
    \label{tab:train_config}
    \vspace{-1.8mm}
    \begin{adjustbox}{max width=1\linewidth}
        \begin{tabular}{l|ccc}
            \toprule
            \bf{Method} & \bf{Optimizer} & \bf{Init. learning rate} & \bf{Learning rate scheduler} \\
            \midrule
            \bf{MoCo v2} & SGD (momentum: 0.9, weight decay: $1.0 \times 10^{-4}$) & 0.03 & cosine anneal (min. lr: 0) \\
            \bf{{\method}-MoCo v2} & SGD (momentum: 0.9, weight decay: $1.0 \times 10^{-4}$) & 0.03 & cosine anneal (min. lr: 0) \\
            \midrule
            \bf{SimSiam} & SGD (momentum: 0.9, weight decay: $1.0 \times 10^{-4}$) & 0.1 & cosine anneal (min. lr: 0) \\
            \bf{{\method}-SimSiam} & SGD (momentum: 0.9, weight decay: $1.0 \times 10^{-4}$) & 0.1 & cosine anneal (min. lr: 0) \\
            \midrule
            \bf{SwAV} & SGD (momentum: 0.9, weight decay: $1.0 \times 10^{-6}$) + LARS~\cite{lars} & 4.8 & 10 epochs warmup (lr: $0.3\rightarrow4.8$) + cosine anneal (min. lr: $4.8 \times 10^{-3}$) \\
            \bf{{\method}-SwAV} & SGD (momentum: 0.9, weight decay: $1.0 \times 10^{-6}$) + LARS~\cite{lars} & 4.8 & 10 epochs warmup (lr: $0.3\rightarrow4.8$) + cosine anneal (min. lr: $4.8 \times 10^{-3}$) \\
            \midrule
            \bf{MoCo v3} & AdamW (betas: [0.9, 0.999], weight decay: 0.1) & $1.6 \times 10^{-3}$ & 40 epochs warmup (lr: $0 \rightarrow 1.6 \times 10^{-3}$) + cosine anneal (min. lr: 0) \\
            \bf{{\method}-MoCo v3} & AdamW (betas: [0.9, 0.999], weight decay: 0.1) & $1.6 \times 10^{-3}$ & 40 epochs warmup (lr: $0 \rightarrow 1.6 \times 10^{-3}$) + cosine anneal (min. lr: 0) \\
            \midrule
            \bf{DINO} & AdamW (betas: [0.9, 0.999], weight decay: $0.04 \rightarrow 0.4$\dag) & $3.0 \times 10^{-3}$ & 10 epochs warmup (lr: $0 \rightarrow 3.0 \times 10^{-3}$) + cosine anneal (min. lr: $2.0 \times 10^{-6}$) \\
            \bf{{\method}-DINO} & AdamW (betas: [0.9, 0.999], weight decay: $0.04 \rightarrow 0.4$\dag) & $3.0 \times 10^{-3}$ & 10 epochs warmup (lr: $0 \rightarrow 3.0 \times 10^{-3}$) + cosine anneal (min. lr: $2.0 \times 10^{-6}$) \\
            \midrule
            \bf{iBOT} & AdamW (betas: [0.9, 0.999], weight decay: $0.04 \rightarrow 0.4$\dag) & $3.0 \times 10^{-3}$ & 10 epochs warmup (lr: $0 \rightarrow 3.0 \times 10^{-3}$) + cosine anneal (min. lr: $2.0 \times 10^{-6}$) \\
            \bf{{\method}-iBOT} & AdamW (betas: [0.9, 0.999], weight decay: $0.04 \rightarrow 0.4$\dag) & $3.0 \times 10^{-3}$ & 10 epochs warmup (lr: $0 \rightarrow 3.0 \times 10^{-3}$) + cosine anneal (min. lr: $2.0 \times 10^{-6}$) \\
            \bottomrule
        \end{tabular}
    \end{adjustbox}
\end{spacing}
\scriptsize{\dag$\,$ Weight decay is not applied on bias and normalization parameters; the weight decay rate increases from 0.04 to 0.4 along training.}
\vspace{0.6mm}
\end{table}

%%%%%%%%%%%%%%%%%%%%%%%%%%%%%%%%%%%%%%%%%%%%%%%%%%%%%%%%%%%%

% optimizer
% learning rate & scheduler

In Tab.~\ref{tab:train_config}, we list the optimizer, initial learning rate and learning rate scheduler for each baseline and its variant under {\method}. In general, we follow the official implementation to set the configurations of each baseline, and the adaptation to the {\method} framework does not change any training configuration, which ensures fair comparison. We state the extra configurations specific to each method as below.
\vspace{-0.3mm}
\begin{itemize}
\item \textbf{MoCo v2} constructs a single positive pair per sample for contrastive learning. Each view of a pair is transformed from the raw image by a sequential of random crop, resize to $224 \times 224$, color transformation and random horizontal flip. We refer readers to the original paper~\cite{moco_v2} for detailed parameters of each transformation function. MoCo v2 maintains a queue of 65,536 image embeddings as negative samples. For a positive pair, the embedding of one of its view is extracted by a vanilla encoder updated by gradients per training step, and the embedding of another view is extracted by a momentum encoder whose parameters are the moving average of the vanilla encoder's parameters, where the momentum is set as 0.999. 
\item \textbf{SimSiam} also uses a single positive pair per sample for cross-view prediction. Its transformation scheme for each view of a pair follows that of MoCo v2. The learning of SimSiam does not rely on negative pairs. 
\item \textbf{SwAV} constructs 2 global views and 6 local views per sample to perform cross-view cluster assignment prediction. Both the global and local views of an image are derived by sequentially applying random crop, resize, color transformation and random horizontal flip. The global view is resized to $224 \times 224$, and the local view is resized to $96 \times 96$. We refer readers to the original paper~\cite{swav} for other transformation parameters.
\item \textbf{MoCo v3} follows MoCo v2 to derive positive pairs. However, instead of maintaining a queue of negative samples as in MoCo v2, it utilizes other samples in a mini-batch to construct negative pairs, which is demonstrated to be better when training with large batch size. MoCo v3 also maintains a momentum encoder along training, where the value of momentum increases from 0.99 to 1 with a cosine rate. 
\item \textbf{DINO} constructs 2 global views and 10 local views per sample for self-distillation across different views. Both the global and local views are derived by a sequential of random crop, resize, random horizontal flip and color transformation. The global view is resized to $224 \times 224$, and the local view is resized to $96 \times 96$. Readers are referred to the original paper~\cite{dino} for other transformation parameters. DINO maintains a teacher network as the moving average of the student network, where the momentum for updating the teacher network increases from 0.996 to 1 along the training process with a cosine rate. Also, DINO uses the momentum update scheme to maintain the center value of each representation dimension, in which the momentum is set as 0.9. 
\item \textbf{iBOT} follows DINO to derive local and global views for self-distillation and masked image modeling. iBOT also maintains a teacher network and center values of image representations following the schemes in DINO. iBOT additionally maintains a center value for each dimension of patch representation by momentum update, where the momentum is set as 0.9.   
\end{itemize}

%%%%%%%%%%%%%%%%%%%%%%%%%%%%%%%%%%%%%%%%%%%%%%%%%%%%%%%%%%%%

\subsection{More Evaluation Details of Transfer Learning} \label{supp_sec1_3}

% CNN detection: methods follow HCSC
% CNN seg: methods follow HCSC, use different prediction head for seg
% ViT detection: use MM detection, use cascaded mask-rcnn (w/ fpn), use the 4th, 6th, 8th, 12th Transformer blocks' output as FPN's multi-granular inputs; following multi-scale training, optimizer: AdamW (betas: [0.9, 0.999], weight decay: 0.05, layer decay: 0.65), learning rate (mocov3: 1.6e-4, dino, ibot: 1e-4), scheduler: 12 eps, decay on 9 and 11 by the factor 0.1
% ViT seg: methods follow ViT detection, use different prediction head for seg

% In general, we follow the configurations of MoCo v2~\cite{moco_v2} to perform object detection and instance segmentation for all CNN based methods, 
We apply the detection and segmentation configurations of MoCo v2~\cite{moco_v2} to all CNN based methods, and the configurations of iBOT~\cite{ibot} are applied to all Vision Transformer based methods. 

\textbf{Object detection for CNN based methods.} For object detection on COCO~\cite{coco}, we employ the Faster-RCNN-C4~\cite{faster_rcnn} as the object detector and initialize it with the model weights pre-trained by different methods. We train the model on the train2017 split of COCO and evaluate on the val2017 split. An SGD optimizer (initial learning rate: 0.02, weight decay: $1.0 \times 10^{-4}$, momentum: 0.9, batch size: 16) is used to train the model for 180,000 iterations, and the learning rate is warmed up for 100 iterations and decayed by a factor of 0.1 at the 120,000th and 160,000th iteration.   

\textbf{Instance segmentation for CNN based methods.} For instance segmentation on COCO, as in Mask R-CNN~\cite{mask_rcnn}, we basically follow the model architecture for object detection and additionally append a mask prediction head for segmentation. All training configurations are identical between the detection task and the segmentation task. 

\textbf{Object detection for Vision Transformer based methods.} We utilize the MMDetection toolbox~\cite{mmdetection} to evaluate various pre-trained Vision Transformers on COCO object detection. The object detector is built under the Cascade Mask R-CNN architecture~\cite{cascade_rcnn}, where the outputs of ViT-B/16~\cite{vit} at the 4th, 6th, 8th and 12th Transformer block serve as the inputs of the Feature Pyramid Network (FPN)~\cite{fpn}. We employ an AdamW optimizer (betas: [0.9, 0.999], weight decay: 0.05, learning rate layer decay factor~\cite{beit}: 0.65, batch size: 16) to train the model for 12 epochs. The initial learning rate is set as $1.6 \times 10^{-4}$ for MoCo v3 and is set as $1.0 \times 10^{-4}$ for DINO and iBOT, and the learning rate is decayed by a factor of 0.1 at the 9th and 11th epoch. We follow the multi-scale training scheme used in iBOT, and the multi-scale test is not applied. 

\textbf{Instance segmentation for Vision Transformer based methods.} Also, for COCO instance segmentation with Vision Transformer, we append a mask prediction head to the Vision Transformer based detection model for segmentation, which follows the design of Mask R-CNN. All training configurations are aligned between detection and segmentation. 

%%%%%%%%%%%%%%%%%%%%%%%%%%%%%%%%%%%%%%%%%%%%%%%%%%%%%%%%%%%%
%%%%%%%%%%%%%%%%%%%%%%%%%%%%%%%%%%%%%%%%%%%%%%%%%%%%%%%%%%%%

\begin{table}[t]
\begin{spacing}{1.15}
    \centering
    \caption{Results of SwAV and {\method}-SwAV under 200/800 epochs training. Downstream tasks: KNN evaluation (Top1-Acc), linear classification (Top1-Acc), fine-tuning (Top1-Acc) and semi-supervised learning with 1\% or 10\% data (Top1-Acc). We use Tesla-V100-32GB GPUs for all experiments.}
    \label{tab:longer_training}
    \vspace{-1.8mm}
    \begin{adjustbox}{max width=1\linewidth}
        \begin{tabular}{l|ccccc|ccc|cc}
            \toprule
            \bf{Method} & \bf{Arch.} & \bf{\# Epochs} & \bf{Batch size} & \bf{\# GPUs} & \bf{Train time} & \bf{KNN} & \bf{Linear} & \bf{Fine-tune} & \bf{Semi-1\%} & \bf{Semi-10\%} \\
            \midrule \midrule
            \bf{SwAV} & ResNet-50 & 200 & 4096 & 32 & 20h 15min & 63.45 & 72.68 & 76.82 & 51.23 & 68.72 \\
            \bf{{\method}-SwAV} & ResNet-50 & 200 & 4096 & 32 & 27h 45min & 63.99 ($\uparrow\!\;$0.54) & 73.43 ($\uparrow\!\;$0.75) & 77.18 ($\uparrow\!\;$0.36) & 52.12 ($\uparrow\!\;$0.89) & 69.32 ($\uparrow\!\;$0.60) \\
            \midrule
            \bf{SwAV} & ResNet-50 & 800 & 4096 & 32 & 81h 2min & 64.84 & 73.36 & 77.77 & 52.72 & 70.15 \\
            \bf{{\method}-SwAV} & ResNet-50 & 800 & 4096 & 32 & 107h 30min & \textbf{65.43} ($\uparrow\!\;$0.59) & \textbf{74.80} ($\uparrow\!\;$1.44)  & \textbf{78.05} ($\uparrow\!\;$0.28) & \textbf{53.92} ($\uparrow\!\;$1.20) & \textbf{70.62} ($\uparrow\!\;$0.47) \\
            \bottomrule
        \end{tabular}
    \end{adjustbox}
\end{spacing}
\vspace{-1.3mm}
\end{table}

%%%%%%%%%%%%%%%%%%%%%%%%%%%%%%%%%%%%%%%%%%%%%%%%%%%%%%%%%%%%

\vspace{-0.8mm}
\section{Results of Longer Training on SwAV} \label{supp_sec_2}
\vspace{-0.6mm}

In the original paper of SwAV~\cite{swav}, researchers achieve performance gain by increasing pre-training epochs. Here, we want to (1) reproduce this phenomenon under our codebase and (2) verify that the superiority of {\method}-SwAV over the original SwAV preserves as the increase of pre-training epochs.

\textbf{Training Setups.} We investigate SwAV and {\method}-SwAV under 200 and 800 pre-training epochs, respectively. Except for the pre-training epochs, we keep all other training configurations invariant (see Sec.~\ref{supp_sec1_2} for detailed configurations of SwAV). Under 200 epochs pre-training, {\method}-SwAV takes the first 20 epochs as the first-stage training (\emph{i.e.}, training with only SwAV's objective) and takes the rest epochs as the second-stage training (\emph{i.e.}, training with both SwAV's objective and the objective for semantic path discrimination). Under 800 epochs pre-training, the first 80 epochs serve as the first stage, and the rest epochs serve as the second stage. 

When pre-trained for 200 epochs, {\method}-SwAV adopts 3-layer hierarchical prototypes with prototype numbers $(M_1, M_2, M_3) = (30000, 10000, 1000)$, and 1000 negative paths are sampled for semantic path discrimination. When pre-trained for 800 epochs, {\method}-SwAV uses the hierarchical prototype structure $(M_1, M_2, M_3) = (30000, 10000, 1000)$ and the negative path sampling size 1000. These hyperparameters are determined by the search scheme described in Sec.~\ref{sec:exp_1_3}. Under both 200 and 800 epochs pre-training, we implement the projection heads $h_l$ ($1 \leqslant l \leqslant L$) as 2-layer MLPs with batch normalization and ReLU activation in between. 

\textbf{Evaluation Setups.} 
% To evaluate the performance of different pre-trained models, 
We evaluate pre-trained models with five standard downstream tasks, \emph{i.e.}, KNN evaluation, linear classification, fine-tuning and semi-supervised learning with 1\% or 10\% data. All these tasks are evaluated on the ImageNet~\cite{imagenet} dataset following respective evaluation protocols.

\textbf{Results.} In Tab.~\ref{tab:longer_training}, we report the performance of SwAV and {\method}-SwAV under 200 and 800 epochs pre-training, respectively. The SwAV model trained for 800 epochs clearly outperforms its 200 epochs counterpart on all five tasks, which aligns with the original paper's finding that longer training benefits the performance of SwAV~\cite{swav}. Under both 200 and 800 epochs pre-training, {\method}-SwAV surpasses the SwAV baseline with a clear margin on all five tasks. These improvements demonstrate that {\method} can boost off-the-shelf image SSL methods under different training length. 

%%%%%%%%%%%%%%%%%%%%%%%%%%%%%%%%%%%%%%%%%%%%%%%%%%%%%%%%%%%%
%%%%%%%%%%%%%%%%%%%%%%%%%%%%%%%%%%%%%%%%%%%%%%%%%%%%%%%%%%%%

\vspace{-1.2mm}
\section{More Visualization Results} \label{supp_sec_3}
\vspace{-1.0mm}

\begin{figure}[t]
\centering
    \includegraphics[width=0.86\linewidth]{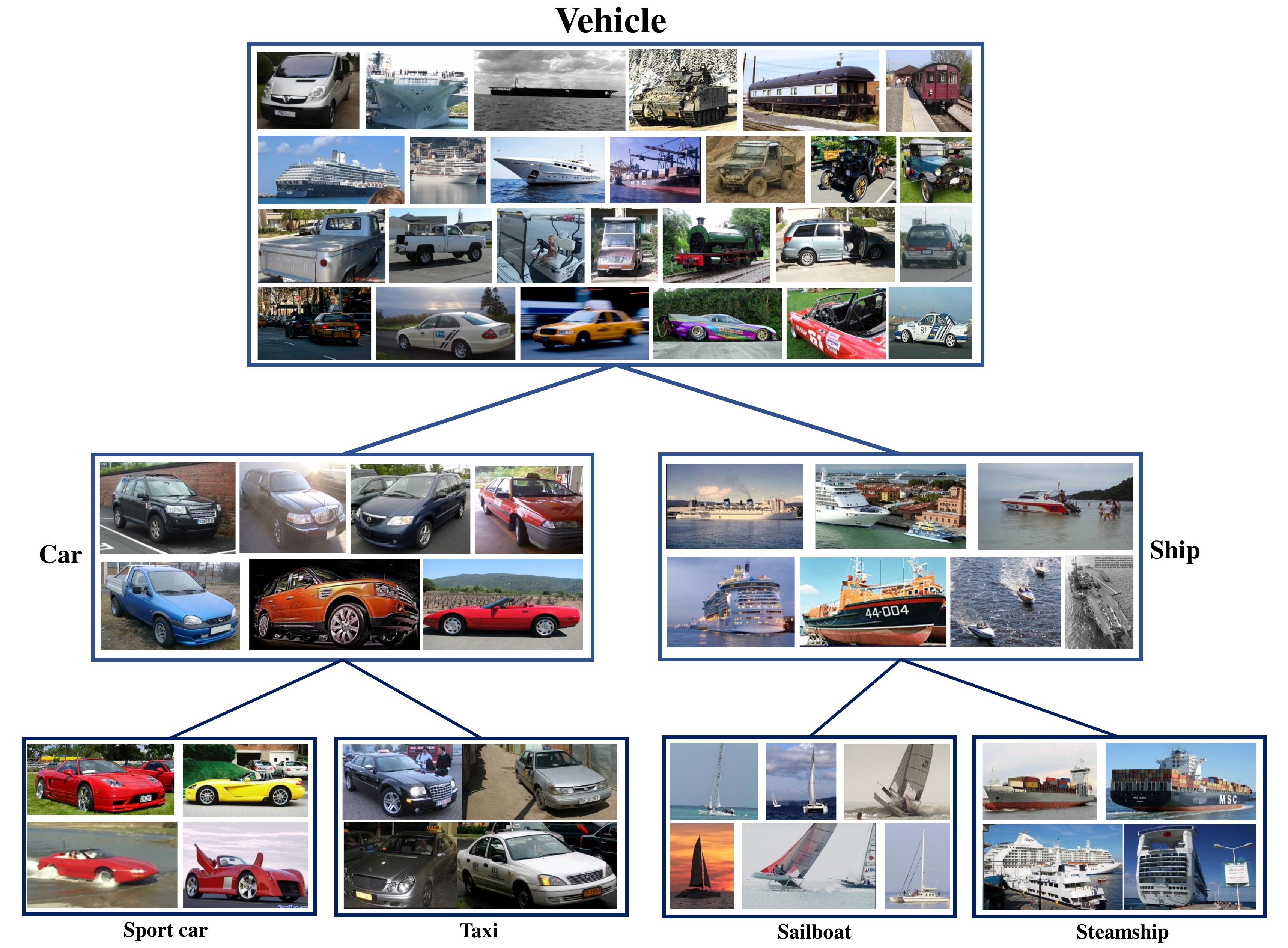}
    \vspace{-2mm}
    \caption{A semantic subtree discovered by {\method}-DINO.}
    \label{fig:tree_1}
    \vspace{-1mm}
\end{figure}

\begin{figure}[t]
\centering
    \includegraphics[width=0.86\linewidth]{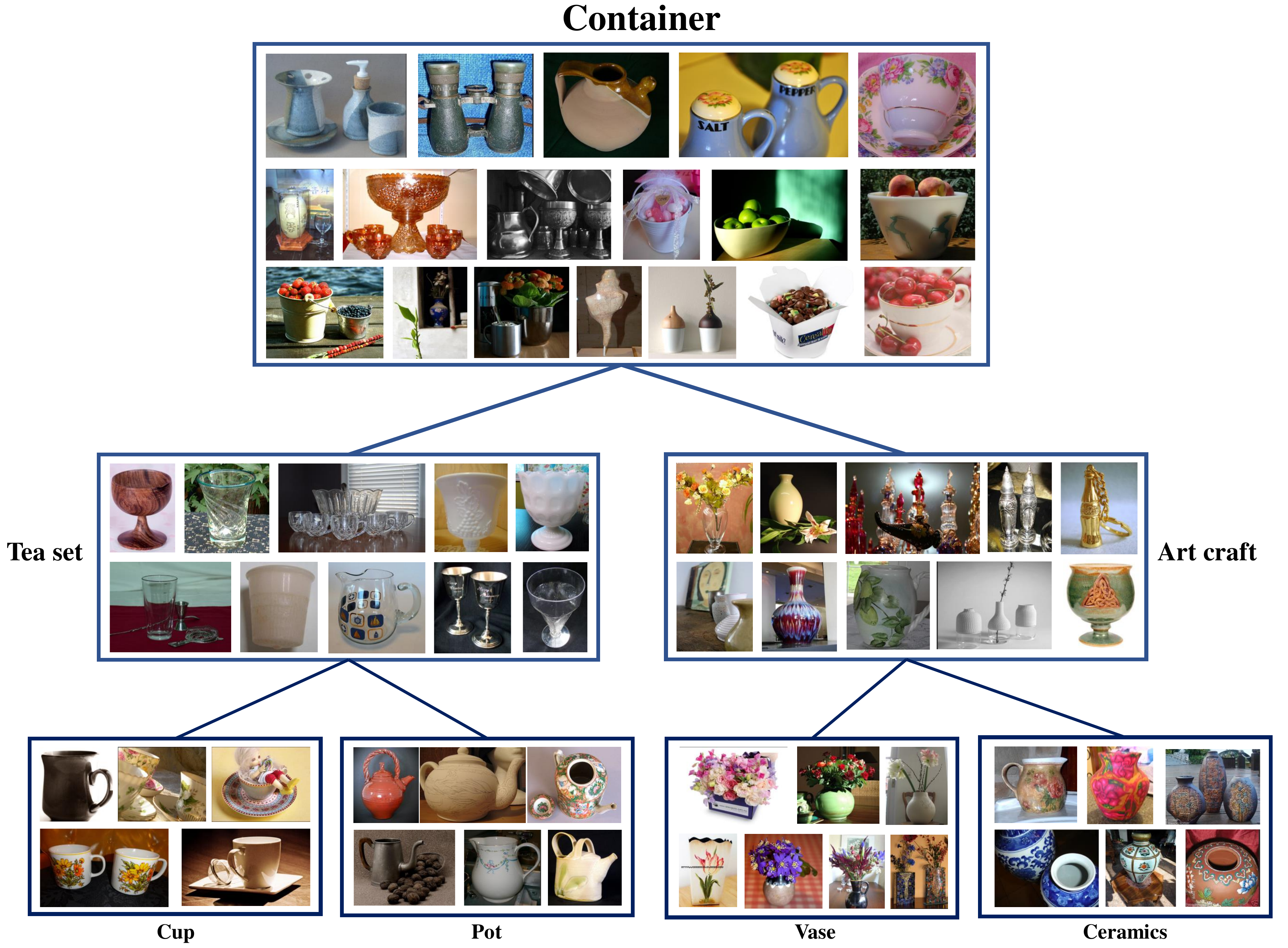}
    \vspace{-2mm}
    \caption{Another semantic subtree discovered by {\method}-DINO.}
    \label{fig:tree_2}
\end{figure}

In Figs.~\ref{fig:tree_1} and \ref{fig:tree_2}, we visualize two typical semantic subtrees within the whole hierarchical prototypes constructed by {\method}-DINO. Each node of a subtree denotes a semantic prototype which is represented by some images assigned to it. The tree in Fig.~\ref{fig:tree_1} discovers the semantic hierarchy of vehicles, \emph{i.e.}, ``vehicle $\rightarrow$ car $\rightarrow$ sport car'', ``vehicle $\rightarrow$ car $\rightarrow$ taxi'', ``vehicle $\rightarrow$ ship $\rightarrow$ sailboat'' and ``vehicle $\rightarrow$ ship $\rightarrow$ steamship''. The tree in Fig.~\ref{fig:tree_2} discovers the semantic hierarchy of containers, \emph{i.e.}, ``container $\rightarrow$ tea set $\rightarrow$ cup'', ``container $\rightarrow$ tea set $\rightarrow$ pot'', ``container $\rightarrow$ art craft $\rightarrow$ vase'' and ``container $\rightarrow$ art craft $\rightarrow$ ceramics''. These results illustrate that, under the proposed {\method} framework, image SSL methods can indeed capture hierarchical semantic information.

%%%%%%%%%%%%%%%%%%%%%%%%%%%%%%%%%%%%%%%%%%%%%%%%%%%%%%%%%%%%
%%%%%%%%%%%%%%%%%%%%%%%%%%%%%%%%%%%%%%%%%%%%%%%%%%%%%%%%%%%%

\vspace{-1.1mm}
\section{Broader Societal Impacts} \label{supp_sec_4}
\vspace{-1.0mm}

This work focuses on designing a general framework to acquire the hierarchical semantic information underlying raw images. This framework learns the model in a totally self-supervised way, which relieves the heavy labor of annotating training samples. Compared to existing self-supervised methods, our approach can acquire more levels of image semantics, which can potentially benefit a broader range of industry applications, like intelligent surveillance under different safety levels. 

However, it cannot be denied that the training process under our framework is time- and resource-intensive. For example, the training of MoCo v2~\cite{moco_v2} takes 19.6 GPU days under our framework, which is more expensive than the 15.1 GPU days training of the original MoCo v2. Therefore, how to train an effective self-supervised image representation model in a more computationally efficient way remains to be further explored. Our future work will seek to address this problem by designing more data-efficient models and training with less data. 

%%%%%%%%%%%%%%%%%%%%%%%%%%%%%%%%%%%%%%%%%%%%%%%%%%%%%%%%%%%%

\newpage

%%%%%%%%%%%%%%%%%%%%%%%%%%%%%%%%%%%%%%%%%%%%%%%%%%%%%%%%%%%%

%%%%%%%%%%%%%%%%%%%%%%%%%%%%%%%%%%%%%%%%%%%%%%%%%%%%%%%%%%%%

\end{document}